\documentclass{bmvc2k}

\newif\ifdebug

\graphicspath {{Images/}}

\usepackage{booktabs}
\usepackage{blindtext}
\usepackage{color}
\usepackage[linesnumbered,lined,ruled,vlined]{algorithm2e}
\SetAlFnt{\scriptsize}

\SetCommentSty{mycommfont}
\usepackage[font=small, skip=0pt]{caption}
\usepackage{epstopdf} 
\usepackage{floatrow}
\usepackage[percent]{overpic}
\usepackage[table]{xcolor}
\usepackage{tabularx}
\usepackage{setspace}
\usepackage{float}
\usepackage{times}
\usepackage{epsfig}
\usepackage{graphicx}
\usepackage{amsmath}
\usepackage{amssymb}
\usepackage{wrapfig}
\usepackage{bibentry}
\usepackage[normalem]{ulem}
\usepackage[symbol]{footmisc}
\renewcommand{\thefootnote}{$\star$}
\newcommand\blfootnote[1]{%
  \begingroup
  \renewcommand\thefootnote{}\footnote{#1}%
  \addtocounter{footnote}{-1}%
  \endgroup
}
\nonstopmode

\title{Object Extent Pooling for \\ Weakly Supervised Single-Shot Localization
\blfootnote{\kern-2.2em \textsuperscript{$\star$} Equal contribution as the first author.}}

\addauthor{Amogh Gudi}{amogh@vicarvision.nl}{12}
\addauthor{Nicolai van Rosmalen\textsuperscript{$\star$}}{nicolai@vicarvision.nl}{12}
\addauthor{Marco Loog}{m.loog@tudelft.nl}{2}
\addauthor{Jan van Gemert}{j.c.vangemert@tudelft.nl}{2}

\addinstitution{
 Vicarious Perception Technologies\\
 Amsterdam, The Netherlands
}
\addinstitution{
 Delft University of Technology\\
 Delft, The Netherlands
}

\runninghead{Gudi, van Rosmalen, Loog, van Gemert}{Object Extent Pooling}

\def\etal{\emph{et al}\bmvaOneDot}
\begin{document}
\maketitle
\begin{abstract}
\label{sec:abstract}
In the face of scarcity in detailed training annotations, the ability to perform object localization tasks in real-time with weak-supervision is very valuable.
However, the computational cost of generating and evaluating region proposals is heavy.
We adapt the concept of Class Activation Maps (CAM)~\cite{Zhou2015} into the very first weakly-supervised 
`single-shot' detector that does not require the use of region proposals. To facilitate this, we propose a novel global pooling technique called Spatial Pyramid Averaged Max (SPAM) pooling for training this CAM-based network for object extent localisation with only weak image-level supervision.
We show this global pooling layer possesses a near ideal flow of gradients for extent localization, that offers a good trade-off between the extremes of max and average pooling.
Our approach only requires a single network pass and uses a fast-backprojection technique, completely omitting any region proposal steps. 
To the best of our knowledge, this is the first approach to do so.
Due to this, we are able to perform inference in real-time at 35fps, which is an order of magnitude faster than all previous weakly supervised object localization frameworks.
\end{abstract}


\section{Introduction}
\label{sec:intro}
Weakly supervised object localization methods~\cite{li2016weakly, bilen2016weakly} can predict a bounding box without requiring bounding boxes at train time. Consequently, such methods are less accurate than fully-supervised methods~\cite{Jifeng2016RFCN, linCVPR17featPyrNets, liuECCV16ssdMultiBoxDetector, redmonCVPR16yolo}: it is acceptable to sacrifice accuracy to reduce expensive human annotation effort at \emph{train time}. Similarly, blazing fast fully supervised single-shot object localization methods such as YOLO~\cite{redmonCVPR16yolo} and SSD~\cite{liuECCV16ssdMultiBoxDetector} make a similar trade-off of running speed versus accuracy at \emph{test time}. More accurate methods~\cite{Jifeng2016RFCN,linCVPR17featPyrNets} are slower and thus exclude real-time embedded applications on a camera, drone or car. In this paper we optimize for speed at train time and at test time: We propose the first weakly supervised single-shot object detector that does not need expensive bounding box annotations during train time and also achieves real-time speed at test time. 

Exciting recent work has shown that object detectors emerge automatically in a CNN trained only on global image labels~\cite{Bency,Oquab2015, Zhou2015}. Such methods convincingly show that a standard global max/average-pooling of convolutional layers retain spatial information that can be exploited to locate discriminative object parts. Consequently, they can predict a point inside the ground truth bounding box with high accuracy. We take inspiration from these works and train only for image classification while exploiting the spatial structure of the convolutional layers. Our work differs in that we do not aim for predicting a single point inside the bounding box, we aim to predict full extent of the object: the bounding box itself.

For predicting the object's extent, we have to decide how object parts are grouped together. Different object instances should be separated while different parts of the same object should be grouped together.  Successful state-of-the-art methods on object localization have therefore incorporated a local grouping step in the form of bounding box proposals~\cite{Jifeng2016RFCN,linCVPR17featPyrNets}. After grouping, it is enough to indicate object presence and the object localization task is simplified to a bounding box classification task. In our work, we use no bounding boxes during training nor box proposals during testing. Instead, we let the CNN do the grouping directly by exploiting the pooling layer.

\begin{figure}
\CenterFloatBoxes
\begin{floatrow}
\killfloatstyle
\floatbox[{\capbeside\thisfloatsetup{capbesideposition={left,center},capbesidewidth=0.5\linewidth, captionskip=0pt}}]{figure}[0.9\FBwidth]
  {
  \includegraphics[trim = 0cm 0.1cm 0cm 0cm, clip = true,width = \linewidth]{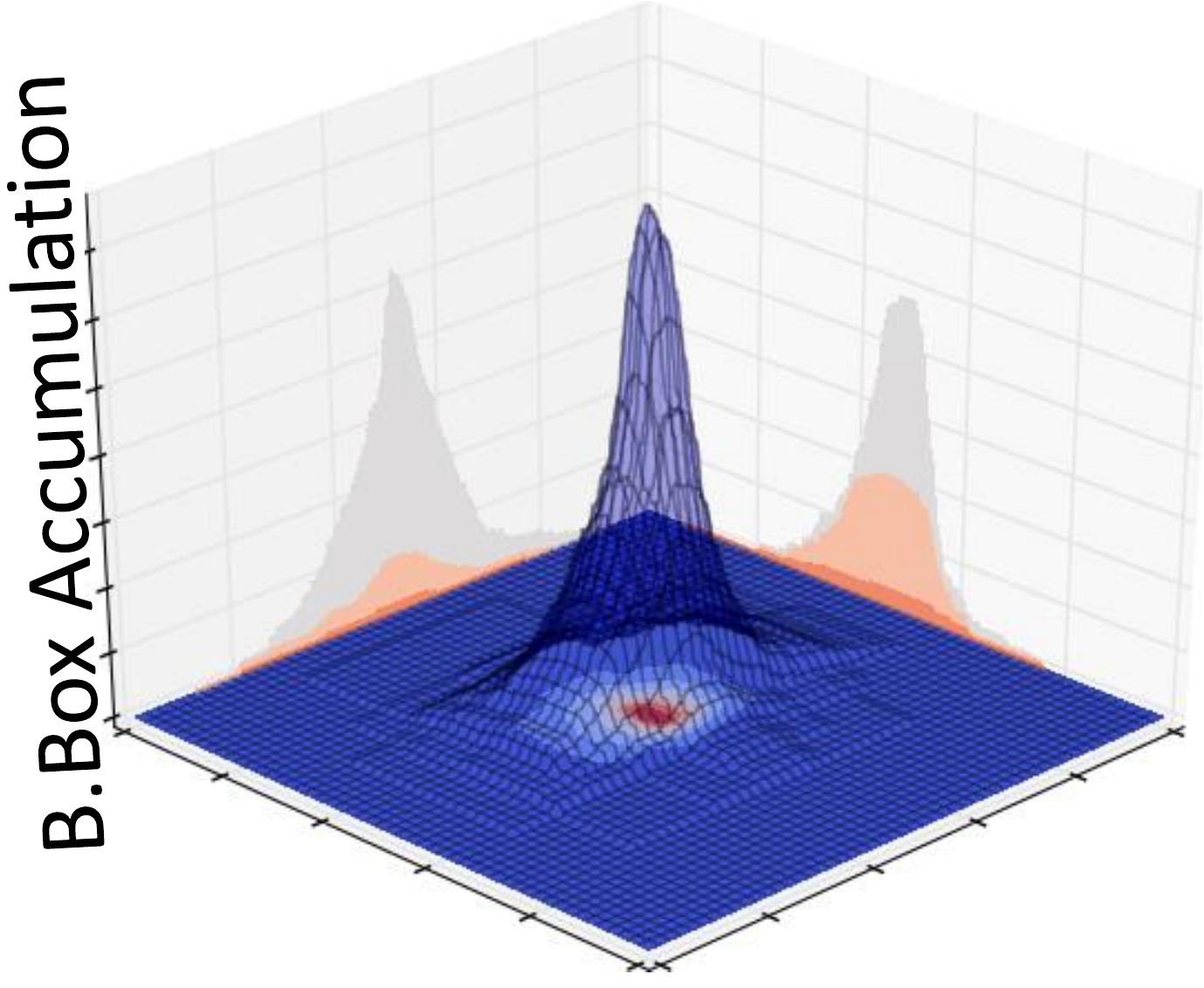}}
  {\captionof{figure}{Accumulation of ground truth bounding boxes of Pascal VOC 2007 centered at the object's maximum activation. Note that the average extent follows a long-tailed distribution.} 
    \label{fig:bbox3d}
  }
\hspace{-5pt}
\floatbox[{\capbeside\thisfloatsetup{capbesideposition={right,center},capbesidewidth=0.5\linewidth}}]{figure}[0.9\FBwidth]
  {\raggedright\includegraphics[trim = 0cm 0.1cm 0cm 0cm, clip = true,width = \linewidth]{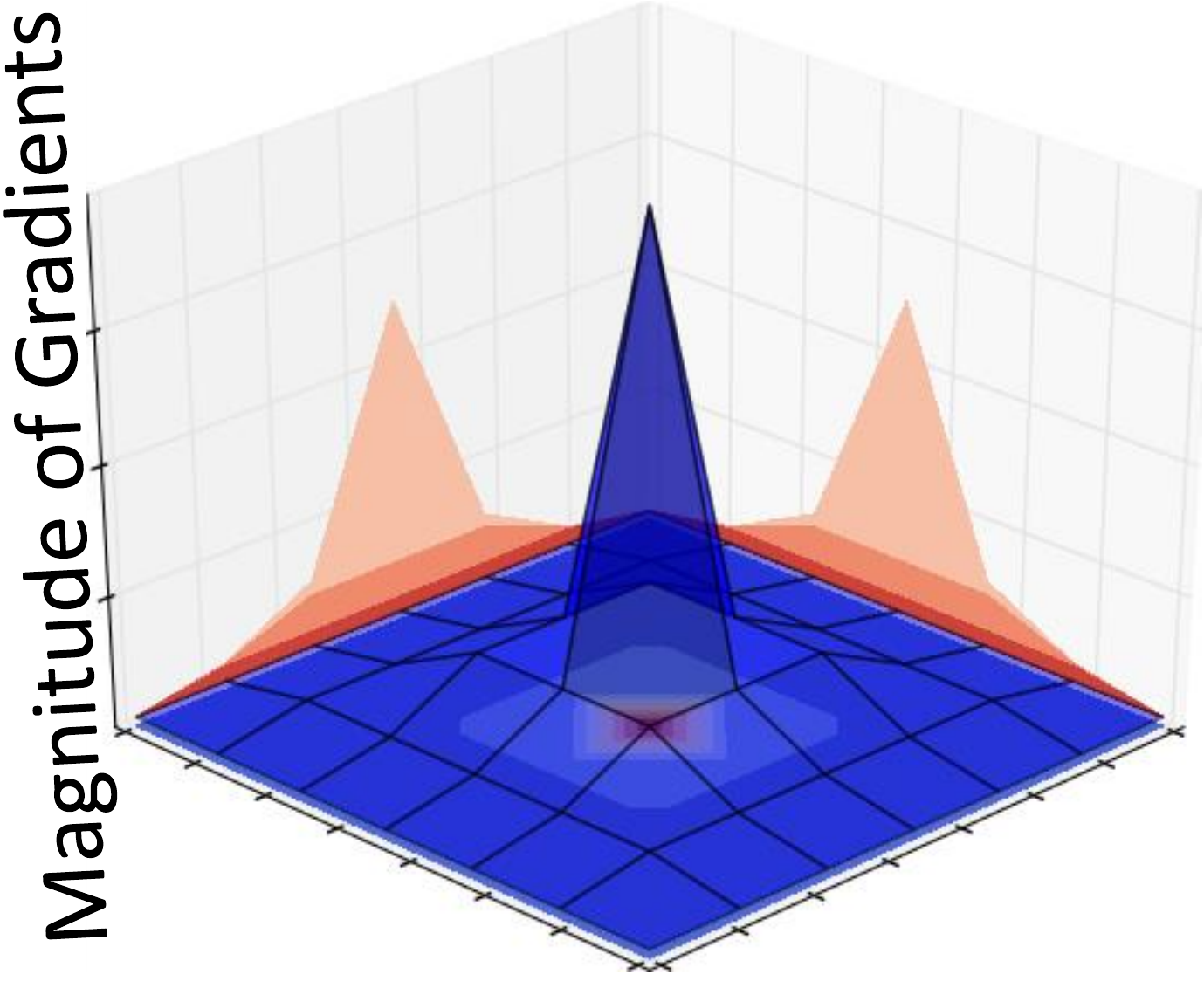}}
  {\captionof{figure}{Gradient flow from our region pooling layer centered around the max activation. Note that our pooling follows the average extent illustrated in Figure \ref{fig:bbox3d}.}
    \label{fig:gradient3d}
  }
\end{floatrow}
\vspace{-15pt}
\end{figure}

The pooling in a CNN groups pixels in a high-resolution image to a lower resolution one. Choices in pooling determine how the gradient is propagated back through the network. In average-pooling, the gradient is shared over all underlying pixels. In the case of a global image label, average-pooling will propagate loss gradients to all pixels in the image equally, which will cover the object but will also cover the background. In contrast, max-pooling only promotes the best point and will thus enforce only a single discriminative object part and not the object extent. Average-pooling is too wide, and max-pooling is too narrow; a regional pooling is needed for retaining the extent. Consider Fig~\ref{fig:bbox3d}, where we center the ground truth bounding boxes around its most discriminative part, given by the maximum filter response~\cite{Oquab2015}. The average object extent is peaked, but has heavy tails. This motivates the need for regional pooling. In Fig~\ref{fig:gradient3d}, we show the gradient flow of our proposed pooling method centered around the maximum response. Our pooling method not only assigns gradients to the maximum or to the full image: it pools regionally.

We present the very first weakly-supervised single-shot detector. It has the following novelties. (i) \textbf{Speed}: we extend the idea of class activation maps (CAM)~\cite{Zhou2015} onto a single stage CNN-only architecture for weakly supervised object localization, that achieves good accuracy while being 10-15 times faster than other related methods. (ii) \textbf{Extent pooling}: a `regional' global pooling technique called the Spatial Pyramid Averaged Max (SPAM) pooling for capturing the object extent from weak image-level labels during training. (iii) \textbf{No region proposals}: We demonstrate a simple and fast back-projection pipeline that avoids the need for costly region proposal algorithms~\cite{Uijlings2013}. This allows our framework to perform real-time inference at 35fps on a GPU.

\section{Related Work}
\label{sec:relwork}
\paragraph{Fully Supervised Object Localization.}
The state of the art is based on the R-CNN \cite{Girshick2014} pipeline which CNN combines the power of a classification network (e.g. ResNet \cite{Technologii2013}) with an SVM classifier and  unsupervised region proposals~\cite{Uijlings2013}. 
This idea was sped up by \cite{Girshick2015} and \cite{Ren2015} and many different algorithms emerged trying to propose the best regions \cite{Alexe2012,Endres2014,Pinheiro2015}, 
including a fully convolutional network~\cite{Long2015} based version called R-FCN \cite{Jifeng2016RFCN}. 
Recently published object detectors \cite{liuECCV16ssdMultiBoxDetector, redmonCVPR16yolo} achieved orders of magnitude faster inference speeds with good accuracies by leaving region-proposals behind and predict bounding boxes in a single-shot. The high speed of our method is borrowed from the single-shot philosophy, albeit without requiring full supervision.

\paragraph{Weak Supervised Object Localization.}
Most methods~\cite{li2016weakly, bilen2016weakly, Wang2014, Cinbis2015} follow a strategy where first, multiple candidate object windows are extracted using unsupervised region proposals~\cite{Uijlings2013}, from each of which feature vector representations are calculated, based on which an image-label trained classifier selects the proper window. 
In contrast, our single-shot method does away with region proposals all together by directly learning the object's extent. 

Li \etal~\cite{li2016weakly} sets the state-of-the-art in this domain. They achieve this by filtering the proposed regions in a class specific way, and using MIL~\cite{Dietterich1997} to classify the filtered proposals. 
Bilen \etal~\cite{bilen2016weakly} achieves similar performance by using an ensemble of two-streamed deep network setup: a region classification stream, and a detection steam that rank proposals.
Wang \etal~\cite{Wang2014} starts with the selective search algorithm to generate region proposals, similar to R-CNN. 
They then use Probabilistic Latent Semantic Analysis (pLSA) \cite{Hofmann1999} to cluster CNN-generated feature vectors into latent categories and create a Bag of Words (BoW) representation to classify proposed regions. The work of Cinbis \etal~\cite{Cinbis2015} uses MIL with region proposals. In our work, we also are weakly-supervised, however, we perform localization in an end-to-end trainable single-pass without using region proposals. 

A recent study by \cite{Oquab2015} follows an alternate approach~\cite{Lin2013} of using global (max) pooling over convolutional activation maps for weakly supervised object localization. 
This was one of the first works to use this approach. Their method gives excellent result for predicting a single point that lies inside an object, while predicting its bounding boxes, via selective search region proposals, yields limited success. In our work, we focus on ascertaining the bounding box extent of the object directly. Further efforts by \cite{Bency} improve upon \cite{Oquab2015} in bounding box extent localization by using a tree search algorithm over bounding boxes derived from all final layer CNN feature maps. In our work, we perform extent localization of an object by filtering CNN activations into a single feature map instead of using a search algorithm, which makes our approach faster and computationally light, achieving high-speed inference. 

Finally, the concept of class activation mappings in \cite{Zhou2015} serves as a precursor to our architecture.
Like us, they make the observation that different global pooling operations influence the activation maps differently. We build upon their work and introduce object extent pooling.  

\section{Method}
\label{sec:meth}
To allow weak supervision training for localization for a convolutional-only neural network, we use a training framework ending in a convolutional layer with a single feature map (per object class). 
This is followed by a global pooling layer, which pools the activation map of the previous layer into a single scalar value, which depends on the pooling method.
This output is finally connected to a two-class softmax cross-entropy loss layer (per class).
This network setup is then trained to perform image classification by predicting the presence/absence of objects of the target class in the image using standard back-propagation using image-level labels. A visualization of this setup is shown in Figure~\ref{fig:PCLMapproach}.

During inference, the global pooling and the softmax loss layers are removed, thereby the single activation map of the added final convolutional layer becomes the output of the network, in the form of an $N \times N$ grid. Due to the flow of backpropagated gradients through the global pooling layer during training, the weights of this convolutional layer get updated such that
the location and shape of the strongly activated areas in its activation map essentially have a one-to-one relation with the location and shape of the pixels occupied by positive class objects in the image. 
At the same time, the intensity of the activation values in this activation map essentially represent the confidence of the network about the presence of the objects at the specific location.
Borrowing notation from \cite{Zhou2015}, we call this single feature-map output activation a Class Activation Map (CAM). 

Consequently, to extract the location of the object in the image, the CAM activations are thresholded and backprojected onto the input image to localize the positive class objects.

\subsection{The Class Activation Map (CAM) Layer}
\label{sec:PCLM}
\begin{figure}[h]	
	\centering
	\includegraphics[trim = 1cm 4.3cm 1cm 5cm, clip = true, width = 0.75\linewidth]{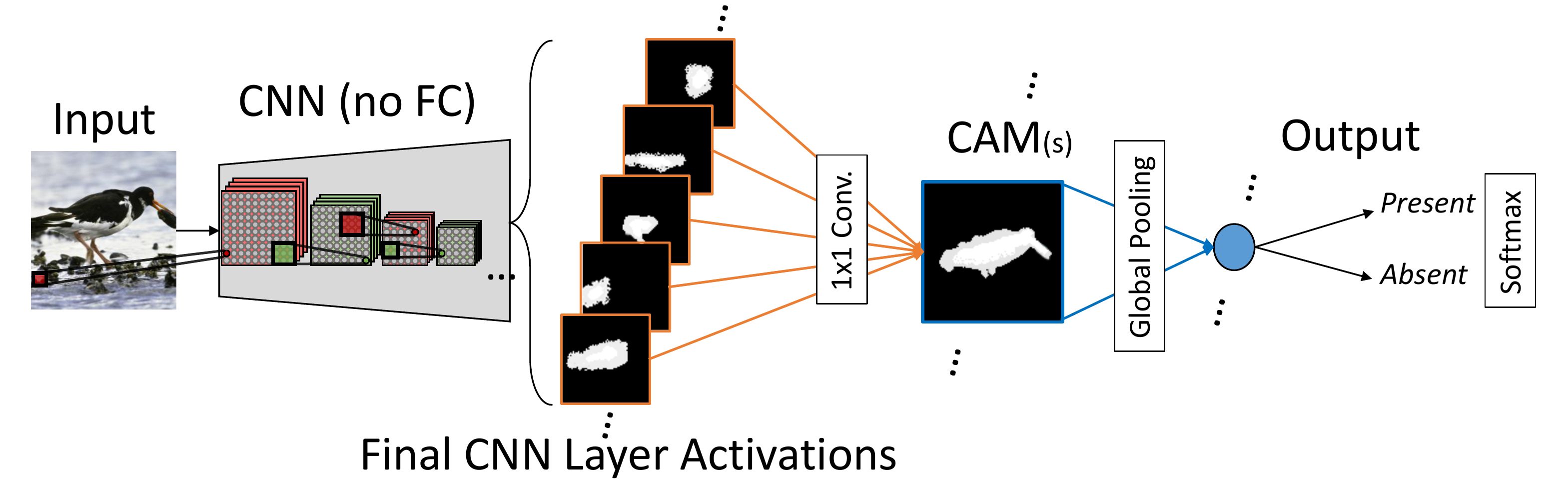}
	\caption{Visualization of the training setup for a CAM-augmented CNN. 
	An extra conv. layer with a single feature map, the CAM, extracts the relevant feature information from the CNN's last conv layer. 
	For weakly supervised training with present/absent annotation, the CAM is followed by a global pooling layer and connected to a softmax output/loss layer.
	}
	\label{fig:PCLMapproach}
\vspace{-7pt}
\end{figure}

The class activation map layer is essentially a simple convolutional layer, albeit with a single feature map/channel (per object class) and a kernel size of $1 \times 1$.
When connected to the final convolutional layer of a CNN, the CAM layer has one separate convolutional weight for each activation map of the previous layer (see Figure \ref{fig:PCLMapproach}).
Training the network under weak-supervision through global pooling and softmax loss updates these kernel weight of the CAM layer through the gradients backpropagated from the global pooling layer. 
Eventually, the feature maps (of the previous conv layer) that produce useful activations for the training task of presence/absence classification are weighted higher, while the feature maps whose outputs are uncorrelated with the presence/absence of the positive class objects are weighted lower. 
Hence, the CAM output can be seen as the weighted sum combination of the activations of all the feature maps of the previous convolutional layer.
Finally after training, the CAM activation essentially forms a heatmap of location likelihood of positive class objects in the input image.

The CAM layer used here is based on the concept of class activation mapping introduced in \cite{Zhou2015}.
While being algorithmically similar, it should be noted that our CAM layer setup is different from the one in \cite{Zhou2015} in the following way:
we perform the global pooling operation \textit{after} the weight multiplication step (via a 1$\times$1 conv.), while \cite{Zhou2015} does this \textit{before} the weight multiplication step (via a FC layer). 
The reason for this difference is to allow greater ease of implementation and lower computational redundancy (requiring pooling on just one feature map).

\begin{figure}
\vspace{-10pt}
\CenterFloatBoxes
\thisfloatsetup{captionskip=3pt}
\begin{floatrow}
\killfloatstyle
\ffigbox[0.45\textwidth]
  {\begin{algorithm}[H]
    \setlength{\textfloatsep}{0pt}
    \setlength{\intextsep}{0pt}
    \KwIn{$\mathbf{[X]}$, $\mathbf{[Y]}$, ${layer_{CAM}}$, $r$ \tcp{activation pixels in CAM layer, the CAM layer, resize ratio}}
    \KwOut{$bpImage$ \tcp{backprojection on input image}}
    \tcc{for each activation pixel in the CAM layer}
    \ForEach{$\{x, y\}$ in $\{\mathbf{[X]}, \mathbf{[Y]}\}$}{ 
    	$x_0 = x_1 \leftarrow x$; $y_0 = y_1 \leftarrow y$; $l \leftarrow layer_{CAM}$ \tcp{init}
    	\tcc{loop through all layers from CAM to input}
    	\While{$l \neq layer_{input}$}{ 
    	\tcc{s, p, k = stride, padding, kernel size}
    		${\{x,y\}}_0 \leftarrow {\{x,y\}}_0 \times s - p$\\
    		${\{x,y\}}_1 \leftarrow {\{x,y\}}_1 \times s - p + k - 1$\\
    		$l \leftarrow layer_{CAM - 1}$ \tcp{Go to next layer}
    	}
    	\tcc{If ratio is provided, correct locations}
    	\If{$r \neq 0$}{
    		${\{x,y\}}_0 \leftarrow {\{x,y\}}_0 + ({\{x,y\}}_1 - {\{x,y\}}_0) \times r \mathbin{/} 2$\\
    		${\{x,y\}}_1 \leftarrow {\{x,y\}}_1 + ({\{x,y\}}_1 - {\{x,y\}}_0) \times r \mathbin{/} 2$
    	}
    	$bpImage[y_0:y_1,x_0:x_1] = 1$ \tcp{fill bpImage}
    }
    \caption{Fast-backprojection} 
    \label{algo:backprojection}
    \end{algorithm}
  }
  {}
\thisfloatsetup{captionskip=3pt}
\ffigbox[0.51\textwidth]
  {\includegraphics[trim = 0cm 34cm 5.5cm 0cm, clip = true, width=\linewidth]{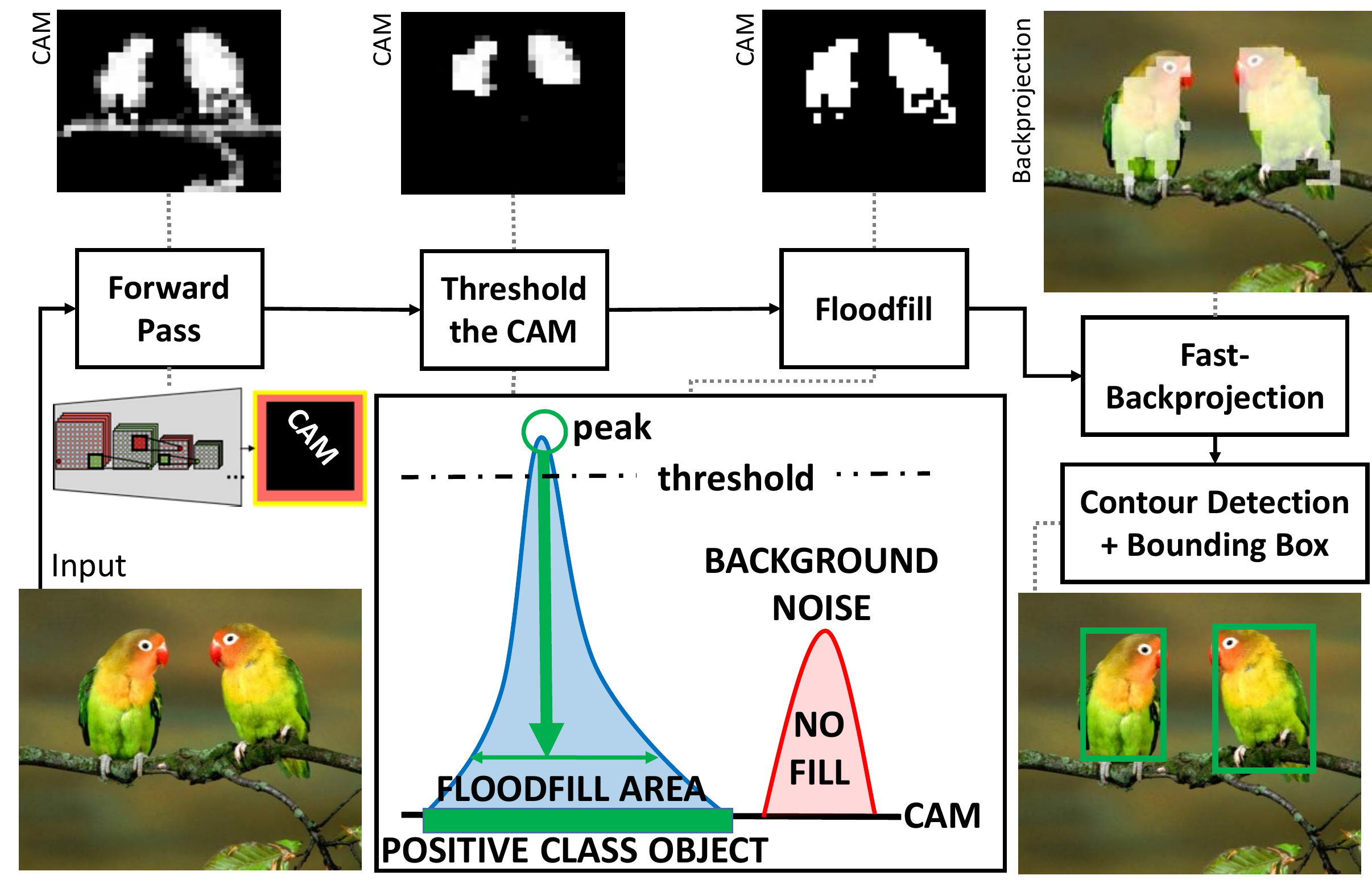}}
  {\captionof{figure}{Visualization of the full inference pipeline. The central plot explains the thresholding and floodfilling steps. The outputs of the pipeline are positive class object bounding boxes.}
    \label{fig:input2output}
  }
\end{floatrow}
\vspace{-18pt}
\end{figure}

\vspace{-15pt}
\subsubsection{Inference}
\label{sec:inference}
The complete pipeline is illustrated in Figure \ref{fig:input2output}. A peak of CAM's activations would occur at the location corresponding to the most discriminative part of the object.
The height of the peak is related to network confidence, whereas the extent of the object is captured by the width. To get a localization proposal, we can investigate which pixels in the original image where responsible for the activations that form a peak in the CAM.
First, only the CAM peaks above the CAM threshold (computed based on the ratio of biases/weights of the output layer, learnt during training) are considered. 
Next, using a floodfill algorithm, all activated pixels belonging to the `mountain' of this peak (including those below the threshold) are selected, as illustrated on the central plot in Figure \ref{fig:input2output}. 
These pixels are then backprojected onto the input image via a fast-backprojection technique explained in Algorithm \ref{algo:backprojection}. 
We call it `fast' because it computes the mapping between CAM pixels and the input pixels without actually performing a backward pass through the network.
As can be inferred, this algorithm backprojects onto all pixels in the input image that could have contributed to the CAM activations (its receptive field). Therefore, we use a ratio parameter $r$ to influence the size of the backprojected area. This parameter can be set by heuristics, or optimised over a separate validation set.
Finally, by performing a contour detection on this backprojection, we can fit simple rectangular bounding boxes on the detected contours to localize the extent of the object.

\subsection{Global Pooling}
\label{sec:global_pooling}
During training, the gradients computed from the loss layer reach the CAM layer through the global pooling layer. 
The connecting weights between the CAM and the previous conv layers are updated based on the distribution/flow of the gradients defined by the type of global pooling layer used.
Hence, the choice of global pooling layer and its distribution of gradients to bottom layers is an important consideration for this framework for weak supervision. 

\vspace{-0.4cm}
\paragraph{Equation Legend} 
In the equations hereafter, we consider a CAM activation map of $N \times N$, where $x_n$ is an arbitrary pixel in it. The backpropagated gradients from the top loss layer is denoted by $g$.

\subsubsection{Max and Average Pooling (GMP \& GAP)}
\label{sec:max_and_average_pooling}
\paragraph{Global Max Pooling (GMP)}
\label{sec:gmp}
layer is essentially a simple max pooling layer commonly used in CNNs, albeit whose kernel size is the same as the input image size. 
During the forward pass, this essentially means it always returns a single scalar pixel whose value is equal to the pixel with the highest value in the input image.
During the backward pass, Equation~\ref{eq:gmp} depicts how the gradients ($\nabla_{GMP}$) are computed for all pixel locations in the CAM layer. 

\begin{wrapfigure}{r}{.461\textwidth}
\vspace{-2\baselineskip}
\begin{equation}
\label{eq:gmp}
  \nabla_{GMP}= g \cdot
  \begin{cases}
    1, & \text{if } x_n = \max\limits_{0 <= n < N} (x_n)\\
    0, & \text{otherwise}
  \end{cases}
\end{equation}
\vspace{-2.1\baselineskip}
\end{wrapfigure}

It can be seen from the equation that the gradient is passed only to the location with the maximum activation in the CAM.
During training with a positive object image, this implies that the detectors that additively contributed in making this pixel value high are encouraged via a positive weight update.
Conversely, for a negative object image, the detectors that contributed in creating the highest value in the CAM are discouraged.
Therefore, the network only learns from the image area that produces max activation in the CAM, i.e., the most discriminative object parts.

\vspace{-0.4cm}
\paragraph{Global Average Pooling (GAP)}
\label{sec:gap}
layer performs a similar global pooling such that the single output pixel is the average of all input pixels during the forward pass.
During the backward pass, the gradients are computed as denoted in Equation \ref{eq:gap}. 

\begin{wrapfigure}{l}{.213\textwidth}
\vspace{-3\baselineskip}
\begin{equation}
\label{eq:gap}
  \nabla_{GAP}= g \cdot \frac{1}{N^2}
\end{equation}
\vspace{-2.5\baselineskip}
\end{wrapfigure}

It can be seen that every location in the CAM gets the same gradient. 
Due to this, over multiple epochs of training, the detectors that fire for parts of the positive class object are strongly weighted, while detectors that fire for everything else are weighted very low.
Thus, the network learns from all input image locations with an equal rate due to GAP's uniform backpropagated gradient.

The visualization of the gradient flow through these pooling layers is shown in Figure \ref{fig:Gradients}.
Due to the single-location max-only gradient distribution of the global max pooling layer, it can be hypothesised that a GMP trained CAM can be quite ideal at pointing to the discriminative parts of an object. 
Conversely, due to the equally spread gradient distribution of the global average pooling layer, a CAM trained with GAP would activate for the full body of object plus parts of correlated or closely situated background. 

\subsubsection{Spatial Pyramid Averaged Max (SPAM) Pooling} 
\label{sec:spam}

Based on the properties of the global max and average pooling layers and from a study of pooling published in \cite{Boureau2010}, 
we propose a pooling layer that is more tuned for training a CAM network for extent localization under weak supervision.

\begin{figure}
\CenterFloatBoxes
\thisfloatsetup{captionskip=0pt}
\begin{floatrow}
\killfloatstyle
\ffigbox[0.45\textwidth]
  {\includegraphics[trim = 0cm 8.3cm 19.4cm 0cm, clip = true, width = \linewidth]{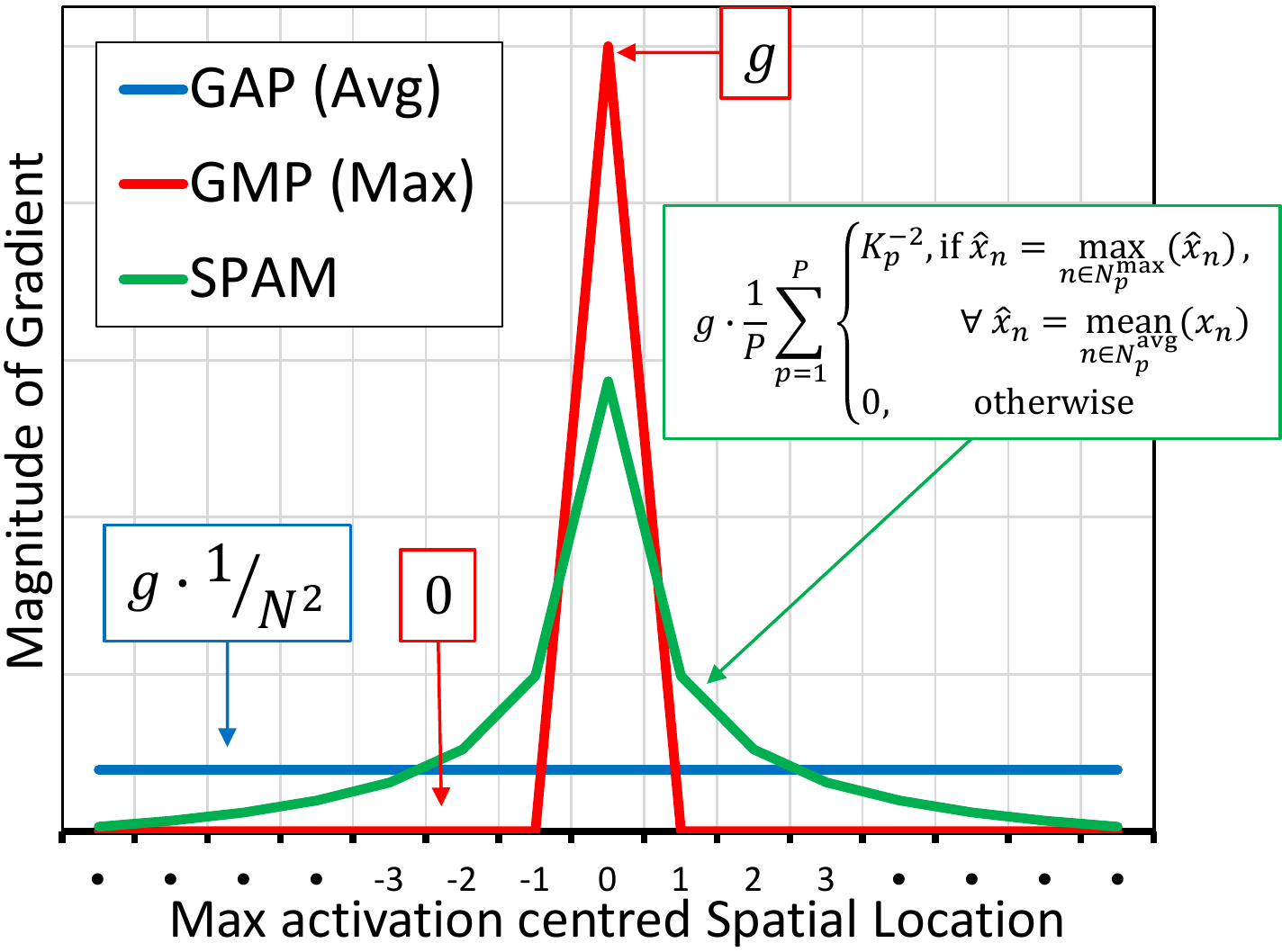}}
  {\captionof{figure}{Visualization of gradient flow through global pooling layers. $g$ is the backpropagated gradient from the upper later. The CAM size considered here is $N \times N$, and centered around its highest activation. SPAM pooling is considered to have $P$ pyramid step, each with an average pooling kernel size of $K_p\times K_p$.}
    \label{fig:Gradients}
  }
\thisfloatsetup{captionskip=0pt}
\ffigbox[0.45\textwidth]
  {\includegraphics[trim = 0.2cm 0.6cm 20.9cm 0.4cm, clip = true, width = 0.7\linewidth, angle =90]{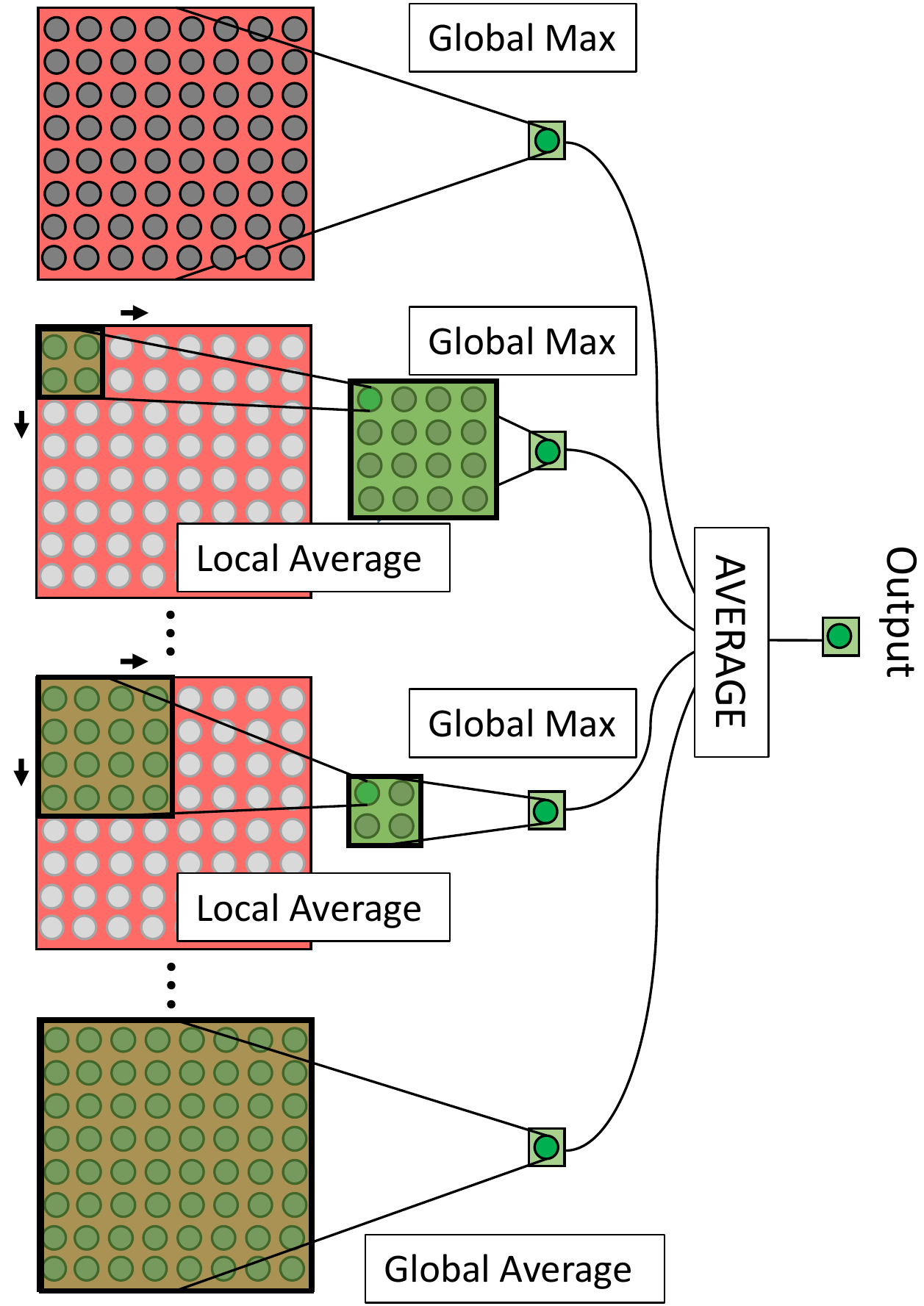}}
  {\captionof{figure}{Architecture of the SPAM layer. First, local average pooling operations are applied in parallel with different kernel sizes, forming a pyramid of output activations. Next, global max pooling is applied and finally, its outputs are averaged. At the ends of the spatial pyramid, we directly show the equivalent GMP and GAP steps.}
    \label{fig:SPGAPvis}
  }
\end{floatrow}
\vspace{-11pt}
\end{figure}

The approach consists of multiple local average pooling operations on the CAM activation map in parallel with varying kernel sizes. 
The kernel size of these average pooling operations is increased in steps (e.g., 1, 2, 4, ...), thus forming a spatial pyramid of local average pooling activation maps.
Next, these activation maps are passed through global max pooling operations, which selects the maximum values among these average pooled activation maps.
Finally, the output single pixel values of these combined pooling operation are averaged together to form the single scalar output of this layer.
Due to the spatial pyramid structure and the use of average and max pooling operations, we call this layer global Spatial Pyramid Averaged Max Pooling, or simply SPAM pooling layer. A visualization of the architecture of SPAM layer is shown in Figure \ref{fig:SPGAPvis}.

During the backward pass, the gradients are computed as depicted in Equation \ref{eq:spam}. Here, we consider a SPAM layer with $P$ pyramid steps, each having a local average pooling kernel size of $K_p\times K_p$; the backpropagated gradients from the top loss layer is represented $g$.

{
\vspace{-0.6\baselineskip}
\begin{equation}
\label{eq:spam}
	\hspace{-0.2cm}
	\nabla_{SPAM} = g \cdot \frac{1}{P} \sum_{p=1}^{P}{
		\begin{cases}
			{K_p}^{-2}, & \text{if } \hat{x}_n = \max\limits_{n \in N^{\text{max}}_p} (\hat{x}_n), 
			\forall \hat{x}_n=\underset{n \in N^{\text{avg}}_p}{\operatorname{mean}}(x_n)\\
		    0, & \text{otherwise}\\
		  \end{cases}
	}
\end{equation}
\vspace{-0.6\baselineskip}
}

{\small \qquad\qquad\qquad where the average/max pool kernel size at pyramid step $p$ is $N^{\text{avg/max}}_p \times N^{\text{avg/max}}_p$.}

The detectors responsible for creating maximal activation receives the strongest update, while the areas surrounding it receive an exponentially lower gradient that is inversely proportional to its distance from the maximal activation.
As a result, while it strongly updates the weights of detectors of discriminative parts responsible for maximal activation, similar to GMP, it still ensures all locations receive a weak update, like in GAP. Due to this property, SPAM layer forms a good middle ground between the extremes of GMP and GAP.
This can also be seen in Figure \ref{fig:Gradients}, which shows the gradients of SPAM layer, in comparison with that of global max and average pooling layers.

The gradient distribution of the SPAM layer is also shown in 3D in Figure \ref{fig:gradient3d}, in comparison with the distribution of ground truth bounding boxes w.r.t the object's most discriminative part (given by CAM's maximal activation).
As can be seen, SPAM's gradients are able to match the distribution of the objects' actual extent.

\section{Experiments and Results}
\label{sec:exp}
\subsection{Evaluation of various Global Pooling strategies on MNIST128}
\paragraph{Setup.}
As a proof of concept, we conduct experiments on a modified MNIST~\cite{LeCun1989} dataset: MNIST128. 
this set consists of 28$\times$28 MNIST digits placed randomly on a blank 128$\times$128 image, thus creating a localization task.
Further, we convert the 10-class MNIST classification problem to a two-class task where the digit 3 (chosen arbitrarily) is considered the positive class, and rest are negative. 
We consider three types of tasks: classification, bounding box localization with at least 0.5 IoU (detection/extent localization), and localization by pin-pointing. Pin-pointing is identifying any single point that falls within the object bounding box~\cite{Oquab2015}. We use a FC-less version of LeNet-5 \cite{LeCun1998} with our CAM extension, trained with softmax loss via various global pooling techniques.
The SPAM pooling layer used here consists of a spatial pyramid of 4 steps, with local average pool kernel sizes $ 1\times 1$, $2 \times 2$, $5 \times 5$, and $N \times N$, where $N$ is the size of the CAM activation map.
After training, the layers succeeding the CAM were removed, and inference was performed as explained in \ref{sec:inference}.

The results of this experiment are in Table \ref{tab:resultsMNIST128NLOC}. 
As hypothesised, GMP is good at locating the most discriminative part of the object, and thus succeeds at pin-pointing, but fails at extent. 
In comparison, GAP performs worse in pin-pointing, and better in extent.
The global SPAM pooling is actually able to perform fairly better overall than both the other forms of pooling for object localisation. 

\begin{table}
\floatbox[{\capbeside\thisfloatsetup{capbesideposition={right,center},capbesidewidth=0.4\linewidth}}]{table}[\FBwidth]
{\caption{Results of the pooling experiments on MNIST128. Bold entries are the ones that perform `well' on the two-class task (\textgreater95 mAP). \label{tab:resultsMNIST128NLOC}}}%
{\scriptsize\begin{tabular}{|lccc|}
\hline
\rowcolor[rgb]{1,0.3,0.3}\cellcolor[gray]{0.8}\textbf{Method} & \multicolumn{3}{c|}{\textbf{mean Average Precision}}\\
\rowcolor[rgb]{1,0.9,1}	& \textbf{Classification} 	& \textbf{Pin-pointing} 		& \textbf{Extent}\\ 
\hline
GMP (Max)	& \multicolumn{1}{|c|}{\textbf{99.8}}	& \multicolumn{1}{|c|}{\textbf{98.9}}	& \multicolumn{1}{|c|}{69.5}\\ 
\rowcolor[rgb]{0.9,0.9,0.9}
GAP (Avg)	& \multicolumn{1}{|c|}{\textbf{99.4}}	& \multicolumn{1}{|c|}{82.3}									& \multicolumn{1}{|c|}{79.1}\\
\rowcolor[rgb]{0.9,0.9,0.9}
SPAM 	& \multicolumn{1}{|c|}{\textbf{99.9}}		& \multicolumn{1}{|c|}{\textbf{95.8}}		& \multicolumn{1}{|c|}{\textbf{95.8}}\\
\hline
\end{tabular}}
\vspace{-10pt}
\end{table}
\vspace{-10pt}
\subsection{Experiments on PASCAL VOC}
\paragraph{Setup}
We adapted an ImageNet pre-trained  version of VGG-16 \cite{Simonyan2014}. We replaced the fully connected layers with our CAM layer, followed by our global SPAM pooling layer plus softmax output layer. 
Once again, the SPAM pooling used here consisted of 4 pyramid steps with kernel sizes of $ 1\times 1$, $2 \times 2$, $5 \times 5$, and $N \times N$, where $N$ is the size of the CAM activation map.
To train our CAM layer weakly on the PASCAL VOC 2007 training set, we assigned a CAM-SPAM-softmax setup, see Fig~\ref{fig:PCLMapproach}, to each of the 20 VOC classes.
After the training, we removed the layers succeeding the CAMs, as was done in the previous experiment.
We also fine-tuned the ratio parameter in Algorithm \ref{algo:backprojection} on a separate validation set. 

\vspace{-3pt}
\subsubsection{Analysis of CAM behaviour trained via various Global Pooling techniques}
\begin{figure}
	\centering
	\floatbox[{\capbeside\thisfloatsetup{capbesideposition={right,center},capbesidewidth=0.4\linewidth}}]{figure}[\FBwidth]
	{\caption{Visualization of the sum of normalized CAM activations, such that the object size present in the image is constant (denoted by the black box). The numbers denote the quantity of activated pixels (correctly) inside vs (wrongly) outside the objects' bounding box.} 
	\label{fig:normlm}}{
		\subfigure[GMP]{\includegraphics[trim = 2cm 2cm 2cm 2cm, clip = true, width=0.3\linewidth]{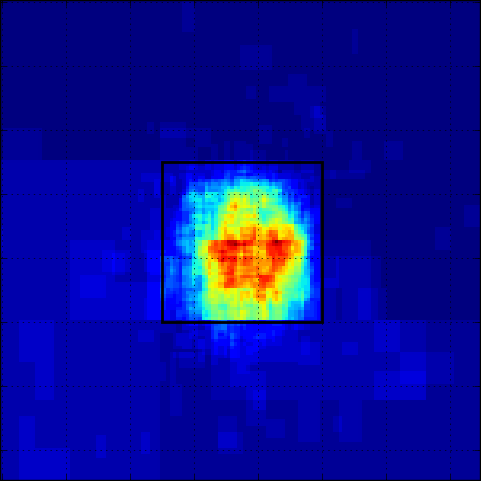}
		\put(-49,62) {\scriptsize inside box: 31K}
		\put(-49,56) {\scriptsize outside box: \textbf{6K}}
		}
		\hspace{1pt}
		\subfigure[SPAM]{\includegraphics[trim = 2cm 2cm 2cm 2cm, clip = true, width=0.3\linewidth]{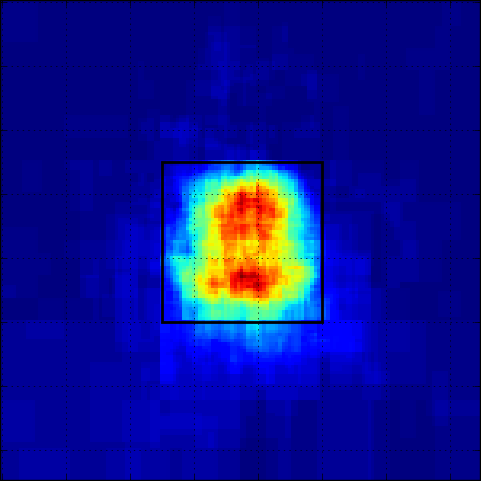}
		\put(-49,62) {\scriptsize inside box: \textbf{88K}}
		\put(-50.5,56) {\scriptsize outside box: \textbf{22K}}
		}
		\hspace{1pt}
		\subfigure[GAP]{\includegraphics[trim = 2cm 2cm 2cm 2cm, clip = true, width=0.3\linewidth]{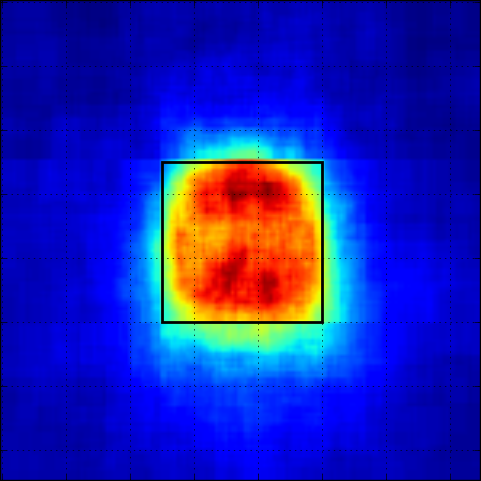}
		\put(-50.5,62) {\scriptsize inside box: \textbf{417K}}
		\put(-52,56) {\scriptsize outside box: 518K}
		}
	}
	 \vspace{-10pt}
\end{figure}
To investigate our method further, we normalize and sum the CAM activations over the whole test set (only images contained one object), such that the size of the object in all the images is constant and centered. In Figure \ref{fig:normlm}, we visualize the distribution of CAM's activated pixels w.r.t the object bounding box.

Figure \ref{fig:normlm} illustrate that the GMP trained CAM activations strongly lie within the bounding region of the object, but fail to activate for the full extent of the object.
Conversely, GAP trained CAM activations spread well beyond the bounds of the object. 
In contrast, the activations of SPAM trained CAM do not spread much beyond the object's boundaries, while still activating for most of the extent of the object.
This observations support our hypothesis that SPAM pooling offers a good trade-off between the adverse properties of GMP and GAP, and hence are better suited for training CAM for weakly supervised localization.

\begin{figure}
	\centering
	\includegraphics[height=.12\textwidth, width=.119\textwidth]{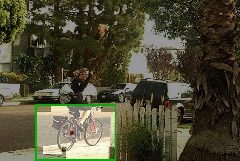} 
	\includegraphics[height=.12\textwidth, width=.12\textwidth]{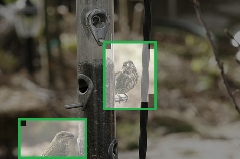} 
	\includegraphics[height=.12\textwidth, width=.119\textwidth]{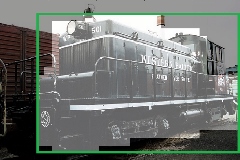} 
	\includegraphics[height=.12\textwidth, width=.12\textwidth]{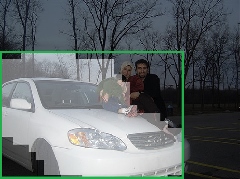} 
	\includegraphics[height=.12\textwidth, width=.12\textwidth]{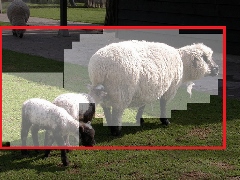} 
	\includegraphics[height=.12\textwidth, width=.119\textwidth]{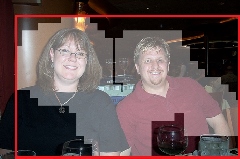} 
	\includegraphics[height=.12\textwidth, width=.12\textwidth]{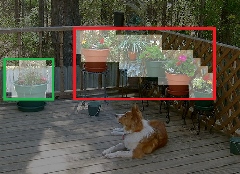} 
	\includegraphics[height=.12\textwidth, width=.119\textwidth]{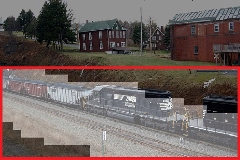} 
	\caption{Localization examples: The highlighted areas in the images indicate the backprojection of CAM activations; green b.boxes match the ground truth, while red do not.
	Note how wrong b.box predictions are mostly either due to closely occurring objects, or closely correlated background.
	}
	\label{fig:PASCALresults}
\end{figure}

\subsubsection{Comparison with the State of the Art}
\begin{figure}
\CenterFloatBoxes
\thisfloatsetup{captionskip=3pt}
\begin{floatrow}
\killfloatstyle
\ffigbox[0.46\textwidth]
  {\footnotesize\setlength\tabcolsep{12pt}\begin{tabular}{|l|c|}
    \hline
    \rowcolor[rgb]{0.88,1,1}
    \cellcolor[gray]{0.8}\textbf{Method}	&	\cellcolor[rgb]{1,0.3,0.3}\textbf{mAP}	\\ \hline
    
    \rowcolor[rgb]{1,0.9,1}\multicolumn{2}{|c|}{\textbf{PASCAL VOC 2007 test set}} \\ \hline
    \rowcolor[rgb]{1,1,0}
    SPAM-CAM\textsuperscript{[Ours]} 	&\textbf{27.5}\\
    \rowcolor[rgb]{0.9,0.9,0.9}
    GMP-CAM (Max Pool)\textsuperscript{[Ours]}	&25.9\\
    GAP-CAM (Avg Pool)\textsuperscript{[Ours]}	&15.6\\
    \hline
    Li\textsuperscript{RP+MIL}~\cite{li2016weakly}	&\textbf{39.5}\\
    \rowcolor[rgb]{0.9,0.9,0.9}
    Bilen\textsuperscript{RP+Ensemble}~\cite{bilen2016weakly}	&39.3\\
    Wang\textsuperscript{RP+pLSA} \cite{Wang2014}	&30.9\\
    \rowcolor[rgb]{0.9,0.9,0.9}
    Cinbis\textsuperscript{RP+MIL} \cite{Cinbis2015}	&30.2\\
    Bency\textsuperscript{RP+TreeSearch}~\cite{Bency}	&25.7\\ 
    \rowcolor[rgb]{1,0.9,1}\multicolumn{2}{|c|}{\textbf{PASCAL VOC 2012 validation set}} \\ \hline
    \rowcolor[rgb]{0.9,0.9,0.9}
    \rowcolor[rgb]{1,1,0}
    SPAM-CAM\textsuperscript{[Ours]} 	&\textbf{25.4}\\
    GMP-CAM (Max Pool)\textsuperscript{[Ours]}	&22.6 \\
    \rowcolor[rgb]{0.9,0.9,0.9}
    GAP-CAM (Avg Pool)\textsuperscript{[Ours]}	&19.3\\ 
    \hline
    Bency\textsuperscript{RP+TreeSearch}~\cite{Bency}	&\textbf{26.5}	\\
    \rowcolor[rgb]{0.9,0.9,0.9}
    Oquab\textsuperscript{RP+GMP} \cite{Oquab2015}	&11.7\\ 
    \hline
    \end{tabular}
  }
  {\captionof{table}{Detection results on PASCAL VOC 2007 \& 2012.
	Entries marked with \textsuperscript{RP} denote their use of region proposal sets.} 
	\label{tab:resultsSOTA}
  }
\thisfloatsetup{captionskip=3pt}
\ffigbox[0.43\textwidth]
  {\includegraphics[trim = 0cm 3.8cm 14.5cm 0cm, clip = true, width = 1.05\linewidth, angle =90]{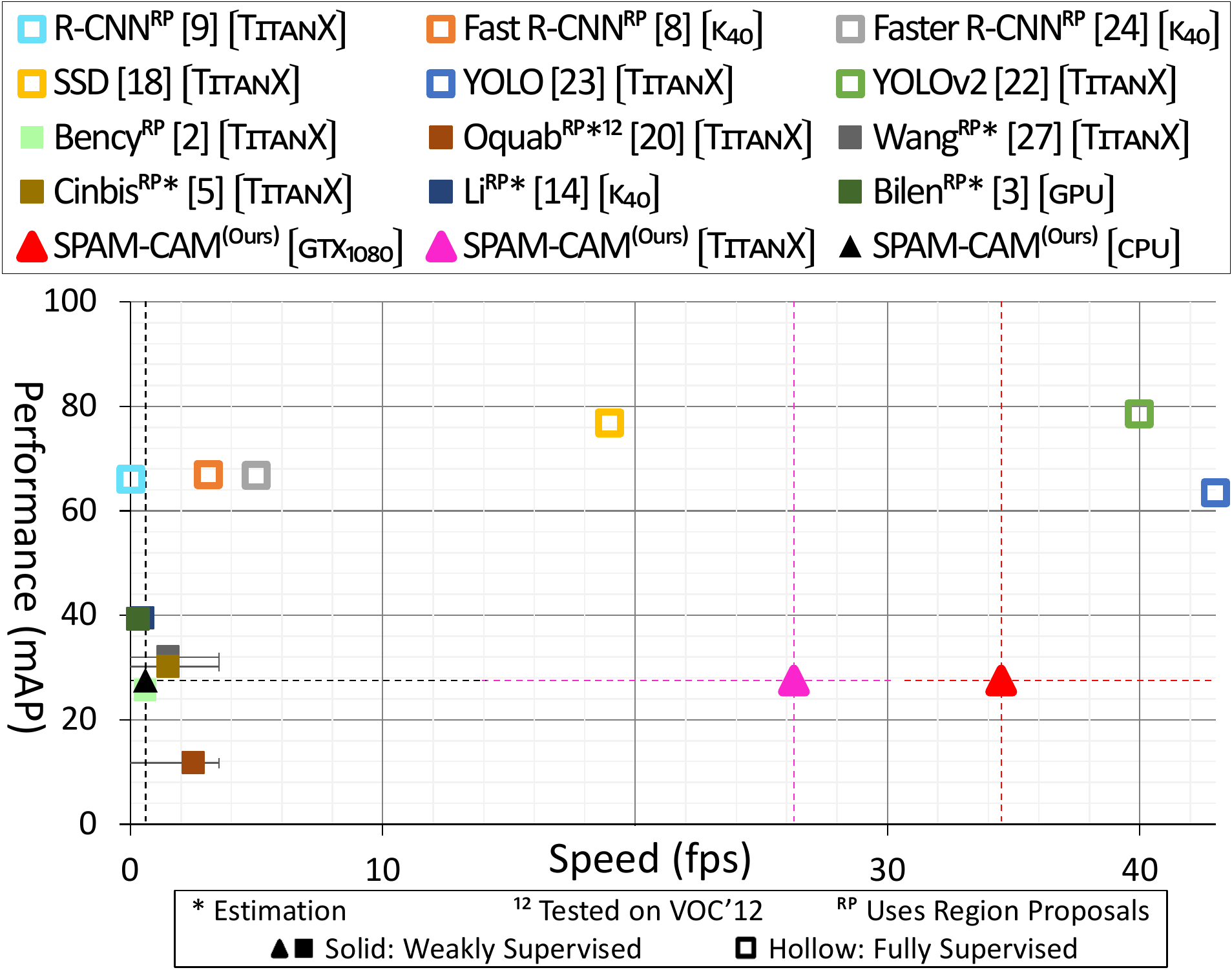}}
  {\captionof{figure}{Speed and performance comparison between different localization methods on PASCAL VOC 2007 test set.}
    		\label{fig:speed}
  }
\end{floatrow}
\end{figure}
The results obtained with this network can be found in Table \ref{tab:resultsSOTA}, in comparison with prior work.
While evaluating these results, it should be noted that all the previous work in this field rely on region proposals, which is an extra computationally heavy step.
\cite{li2016weakly} uses a combination of region proposals, multiple instance learning and fine-tuned deepnets, and \cite{bilen2016weakly} uses region proposals and an ensemble of three deep networks to achieve this performance.
In contrast, our method is purely single-shot, i.e., it requires a single forward pass of the whole image without the need of region proposals, which makes the method computationally very light. 
To the best of our knowledge, this is the first method to perform WSOL without region proposals.

Here, we see that the best methods \cite{li2016weakly,bilen2016weakly} using proposals perform significantly better. However, we are able to match the performance of other methods that also use region proposals \cite{Wang2014,Cinbis2015,Bency,Oquab2015} and rely on similarly sized CNNs as ours. This observation suggests that region proposals themselves are not vital for the task of weakly supervised localization.

\paragraph{Speed Comparison}
\label{sub:speed_comparison}
In Figure~\ref{fig:speed}, the performance of several methods is shown against the speed at which they can achieve this performance (on the PASCAL VOC 2007 test set). 
The test speeds for all methods have been obtained on roughly \textasciitilde500$\times$500 sized images using their default number of proposals, as reported in their respective papers.
Because some studies (\cite{Cinbis2015,Oquab2015,Wang2014}) do not provide details on processing time, we make an estimation based on details of their approach (denoted by *).
In the figure, we also include information on some well known fully-supervised R-CNN approaches \cite{Girshick2014, Girshick2015, Ren2015, liuECCV16ssdMultiBoxDetector, redmonCVPR16yolo, redmon2016yolo9000} for reference.
As can be seen, the VGG-16 based SPAM-CAM performs about 10-15 times faster than all other weakly supervised approaches. 
In fact, even a CPU-only implementation of our approach roughly performs in the same speed range as other TitanX/K40 GPU based implementations.
Additionally, we are able to match the speeds of existing fully supervised single-shot methods like \cite{liuECCV16ssdMultiBoxDetector, redmonCVPR16yolo, redmon2016yolo9000}.

\section{Conclusion}
\label{sec:conc}
In this paper, a convolutional-only single-stage architecture extension based on Class Activation Maps (CAM) is demonstrated for the task of weakly supervised object localisation in real-time without the use of region proposals. 
Concurrently, a novel global Spatial Pyramid Averaged Max (SPAM) pooling technique is introduced that is used for training such a CAM augmented deep network for localising objects in an image using only weak image-level (presence/absence) supervision.
This SPAM pooling layer is shown to posses a suitable flow of backpropagating gradients during weakly supervised training. 
This forms a good middle ground between the strong single-point gradient flow of global max pooling and the equal spread gradient flow of global average pooling for ascertaining the extent of the object in the image.
Due to this, the proposed approach requires only a single forward pass through the network, and utilises a fast-backprojection algorithm to provide bounding boxes for an object without any costly region proposal steps, resulting in real-time inference.
The method is validated on the PASCAL VOC datasets and is shown to produce good accuracy, while being able to perform inference at 35fps, which is 10--15 times faster than all other related frameworks.

\bibliography{../references}
\clearpage
\section*{Appendix}
\begin{figure}[H]
\centering
	\includegraphics[trim = 0cm 13.7cm 0cm 0cm, clip = true, width = \textwidth]{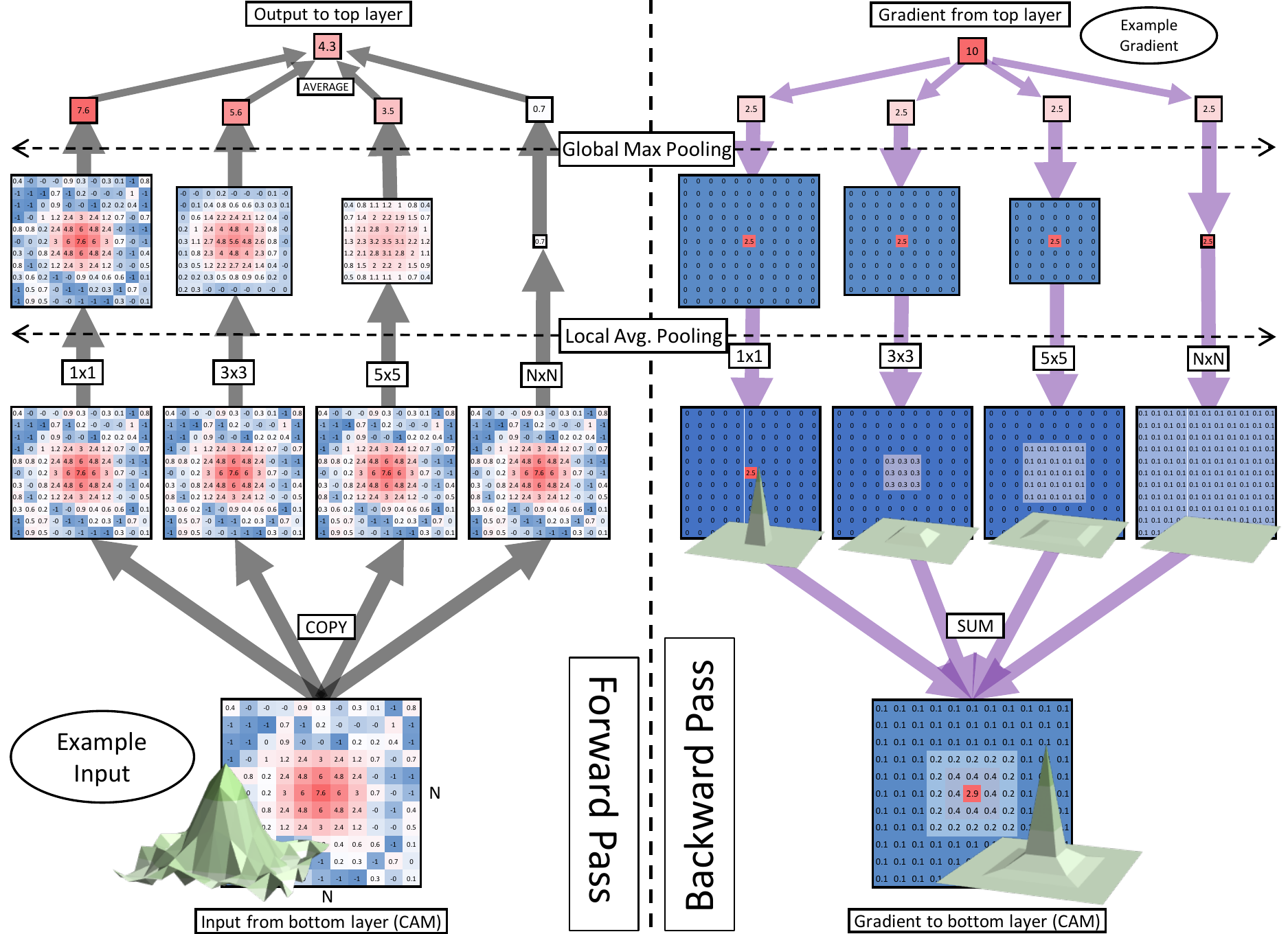}
	\caption{\textbf{Illustration of SPAM's forward and backward pass during training.} 
	During the forward pass, the CAM layer's activations serve as the input to the SPAM pooling layer (bottom left). 
	These activations pass through the pyramid of local average pooling and global max pooling as part of the layer's forward pass. 
	Note that the operations of the first and last pyramid steps with 1x1 and NxN average pooling kernels resemble that of global max pooling and global average pooling layers, respectively. 
	Similarly, during the backward pass, that the gradients (shown in 3D) of the pyramid steps with 1x1 and NxN average pooling kernels are effectively the same as those of global max pooling and global average pooling, respectively.}
	\label{fig:spamdetailed}
\end{figure}

\begin{figure}
\centering
	\includegraphics[trim = 0cm 15cm 27.55cm 0cm, clip = true, width = 0.6\linewidth]{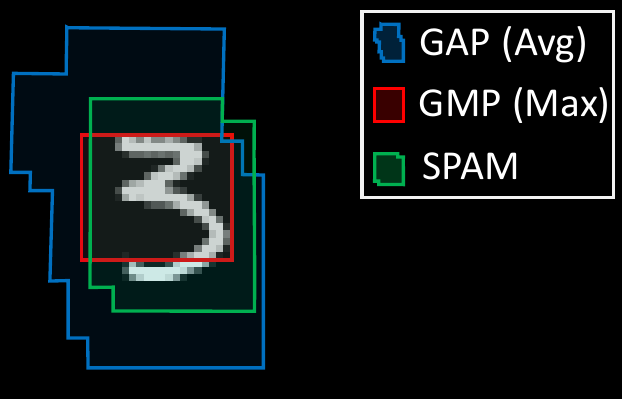}
	\caption{\textbf{Effect of global pooling on backprojection.} This figure shows the training effect of the three global pooling types (global max, global average and global SPAM) on the backprojection of the CAM activations of a LeNet-5 based network. The network was trained on the MNIST128 dataset to classify digit 3 as the positive class. On this typical example image, it can be seen that the backprojection area of a GAP trained CAM is very large, while a CAM trained with GMP backprojects onto a too-small area, likely containing the most discriminative part of the object. The SPAM trained CAM's backprojection more closely aligns with the true boundaries of the positive class object.}
	\label{fig:bb}
\end{figure}

\begin{figure}
\centering
	\includegraphics[width = 0.495\textwidth]{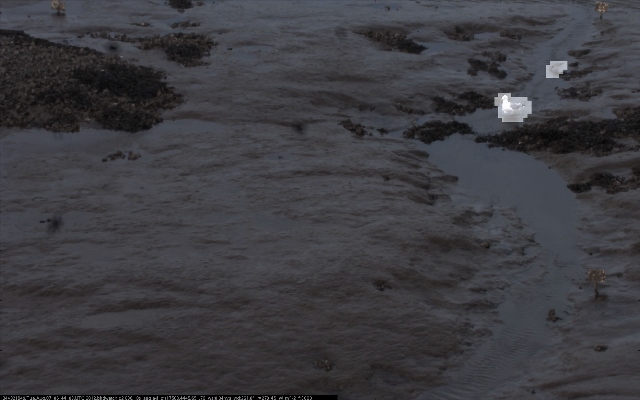}
	\vspace{0.04cm}
	\includegraphics[width = 0.495\textwidth]{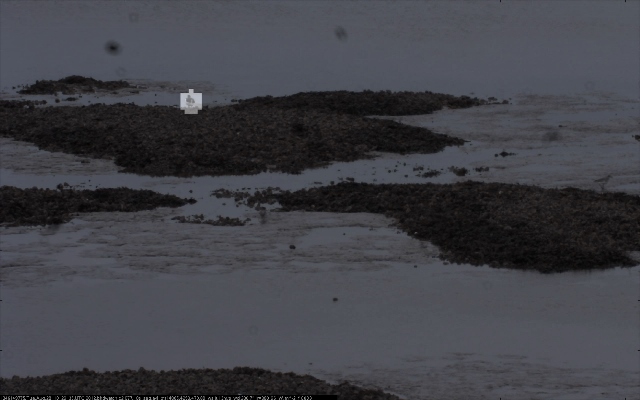}
	\vspace{0.04cm}
	\includegraphics[width = 0.495\textwidth]{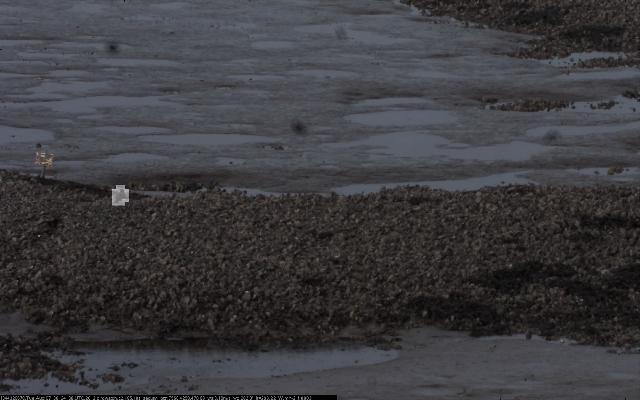}
	\includegraphics[width = 0.495\textwidth]{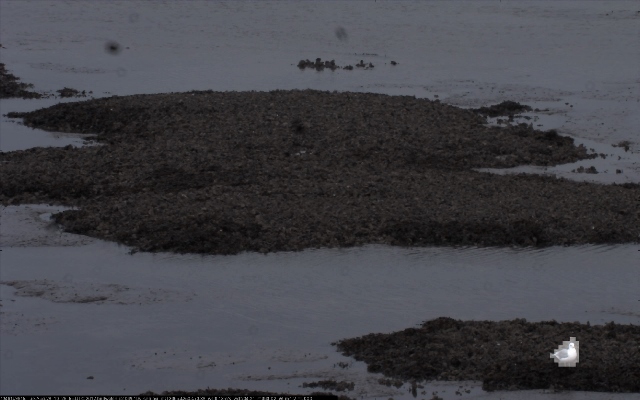}
	\includegraphics[width = 0.995\textwidth]{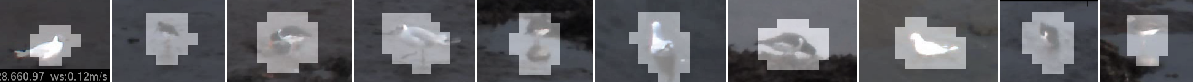}
	\caption{\textbf{Examples of bird localization} on a weakly labelled dataset of CCTV images from a nature reserve. The network used was a SPAM-pooling trained CAM network (VGG-16 based). The bottom row shows the magnified version of the localized birds in additional images.}
	\label{fig:sovonexamples}
\end{figure}

\begin{figure}
\centering
	\includegraphics[width = 0.4\textwidth, height = 0.15\textheight]{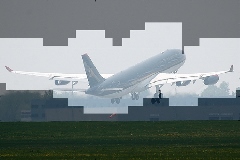}
	\includegraphics[width = 0.4\textwidth, height = 0.15\textheight]{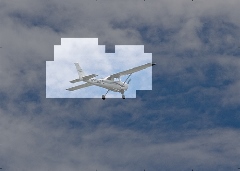}
	\caption{\textbf{Aeroplane} class localization examples from the PASCAL VOC dataset (test set).}
	\label{fig:air}
\end{figure}

\begin{figure}
\centering
	\includegraphics[width = 0.4\textwidth, height = 0.15\textheight]{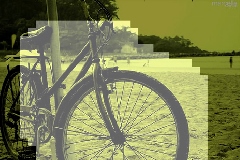}
	\includegraphics[width = 0.4\textwidth, height = 0.15\textheight]{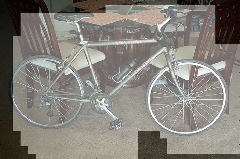}
	\caption{\textbf{Bicycle} (bike) class localization examples from the PASCAL VOC dataset (test set).}
	\label{fig:bike}
\end{figure}

\begin{figure}
\centering
	\includegraphics[width = 0.4\textwidth, height = 0.15\textheight]{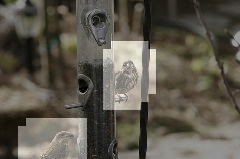}
	\includegraphics[width = 0.4\textwidth, height = 0.15\textheight]{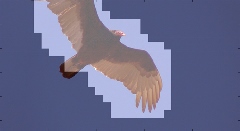}
	\caption{\textbf{Bird} class localization examples from the PASCAL VOC dataset (test set).}
	\label{fig:bird}
\end{figure}

\begin{figure}
\centering
	\includegraphics[width = 0.4\textwidth, height = 0.15\textheight]{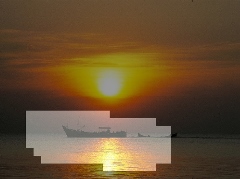}
	\includegraphics[width = 0.4\textwidth, height = 0.15\textheight]{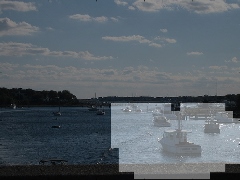}
	\caption{\textbf{Boat} class localization examples from the PASCAL VOC dataset (test set).}
	\label{fig:boat}
\end{figure}

\begin{figure}
\centering
	\includegraphics[width = 0.4\textwidth, height = 0.15\textheight]{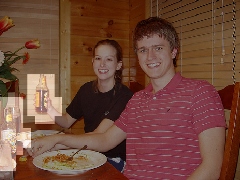}
	\includegraphics[width = 0.4\textwidth, height = 0.15\textheight]{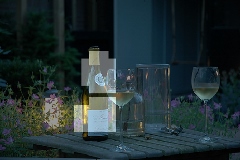}
	\caption{\textbf{Bottle} class localization examples from the PASCAL VOC dataset (test set).}
	\label{fig:bottle}
\end{figure}

\begin{figure}
\centering
	\includegraphics[width = 0.4\textwidth, height = 0.15\textheight]{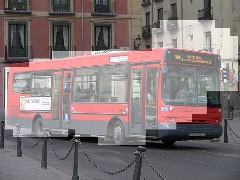}
	\includegraphics[width = 0.4\textwidth, height = 0.15\textheight]{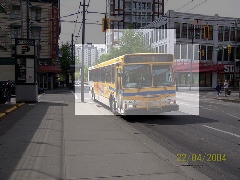}
	\caption{\textbf{Bus} class localization examples from the PASCAL VOC dataset (test set).}
	\label{fig:bus}
\end{figure}

\begin{figure}
\centering
	\includegraphics[width = 0.4\textwidth, height = 0.15\textheight]{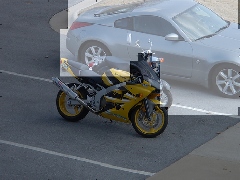}
	\includegraphics[width = 0.4\textwidth, height = 0.15\textheight]{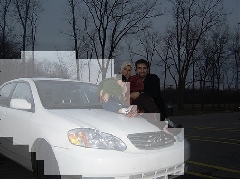}
	\caption{\textbf{Car} class localization examples from the PASCAL VOC dataset (test set).}
	\label{fig:car}
\end{figure}

\begin{figure}
\centering
	\includegraphics[width = 0.4\textwidth, height = 0.15\textheight]{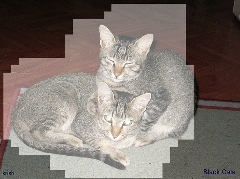}
	\includegraphics[width = 0.4\textwidth, height = 0.15\textheight]{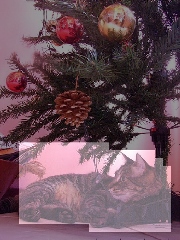}
	\caption{\textbf{Cat} class localization examples from the PASCAL VOC dataset (test set).}
	\label{fig:cat}
\end{figure}

\begin{figure}
\centering
	\includegraphics[width = 0.4\textwidth, height = 0.15\textheight]{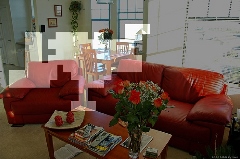}
	\includegraphics[width = 0.4\textwidth, height = 0.15\textheight]{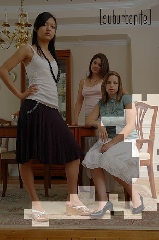}
	\caption{\textbf{Chair} class localization examples from the PASCAL VOC dataset (test set).}
	\label{fig:chair}
\end{figure}

\begin{figure}
\centering
	\includegraphics[width = 0.4\textwidth, height = 0.15\textheight]{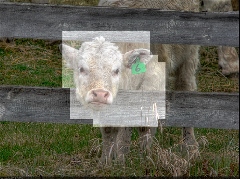}
	\includegraphics[width = 0.4\textwidth, height = 0.15\textheight]{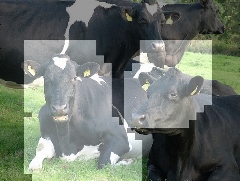}
	\caption{\textbf{Cow} class localization examples from the PASCAL VOC dataset (test set).}
	\label{fig:cow}
\end{figure}

\begin{figure}
\centering
	\includegraphics[width = 0.4\textwidth, height = 0.15\textheight]{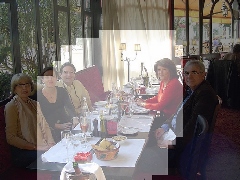}
	\includegraphics[width = 0.4\textwidth, height = 0.15\textheight]{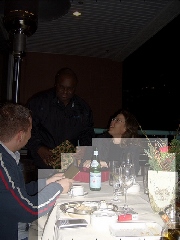}
	\caption{\textbf{Table} class localization examples from the PASCAL VOC dataset (test set).}
	\label{fig:table}
\end{figure}

\begin{figure}
\centering
	\includegraphics[width = 0.4\textwidth, height = 0.15\textheight]{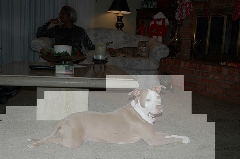}
	\includegraphics[width = 0.4\textwidth, height = 0.15\textheight]{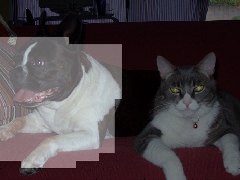}
	\caption{\textbf{Dog} class localization examples from the PASCAL VOC dataset (test set).}
	\label{fig:dog}
\end{figure}

\begin{figure}
\centering
	\includegraphics[width = 0.4\textwidth, height = 0.15\textheight]{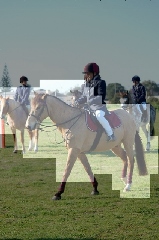}
	\includegraphics[width = 0.4\textwidth, height = 0.15\textheight]{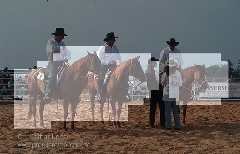}
	\caption{\textbf{Horse} class localization examples from the PASCAL VOC dataset (test set).}
	\label{fig:horse}
\end{figure}

\begin{figure}
\centering
	\includegraphics[width = 0.4\textwidth, height = 0.15\textheight]{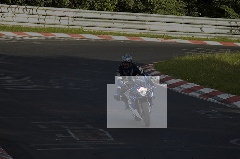}
	\includegraphics[width = 0.4\textwidth, height = 0.15\textheight]{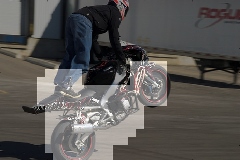}
	\caption{\textbf{Motorcycle} class localization examples from the PASCAL VOC dataset (test set).}
	\label{fig:motorcycle}
\end{figure}

\begin{figure}
\centering
	\includegraphics[width = 0.4\textwidth, height = 0.15\textheight]{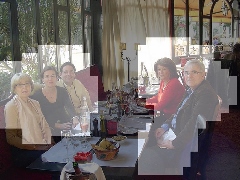}
	\includegraphics[width = 0.4\textwidth, height = 0.15\textheight]{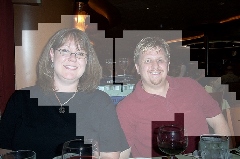}
	\caption{\textbf{Person} class localization examples from the PASCAL VOC dataset (test set).}
	\label{fig:person}
\end{figure}

\begin{figure}
\centering
	\includegraphics[width = 0.4\textwidth, height = 0.15\textheight]{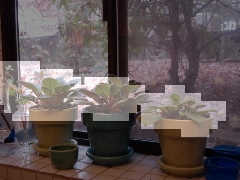}
	\includegraphics[width = 0.4\textwidth, height = 0.15\textheight]{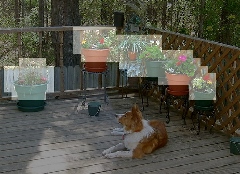}
	\caption{\textbf{Potted Plant} class localization examples from the PASCAL VOC dataset (test set).}
	\label{fig:plant}
\end{figure}

\begin{figure}
\centering
	\includegraphics[width = 0.4\textwidth, height = 0.15\textheight]{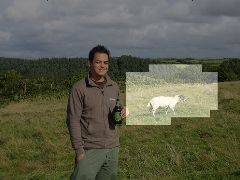}
	\includegraphics[width = 0.4\textwidth, height = 0.15\textheight]{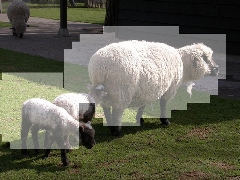}
	\caption{\textbf{Sheep} class localization examples from the PASCAL VOC dataset (test set).}
	\label{fig:sheep}
\end{figure}

\begin{figure}
\centering
	\includegraphics[width = 0.4\textwidth, height = 0.15\textheight]{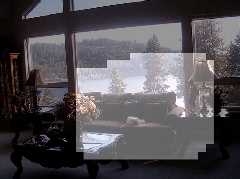}
	\includegraphics[width = 0.4\textwidth, height = 0.15\textheight]{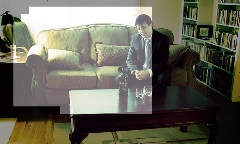}
	\caption{\textbf{Sofa} class localization examples from the PASCAL VOC dataset (test set).}
	\label{fig:sofa}
\end{figure}

\begin{figure}
\centering
	\includegraphics[width = 0.4\textwidth, height = 0.15\textheight]{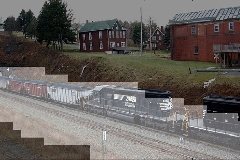}
	\includegraphics[width = 0.4\textwidth, height = 0.15\textheight]{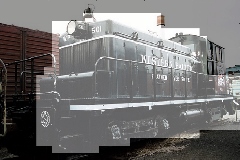}
	\caption{\textbf{Train} class localization examples from the PASCAL VOC dataset (test set).}
	\label{fig:train}
\end{figure}

\begin{figure}
\centering
	\includegraphics[width = 0.4\textwidth, height = 0.15\textheight]{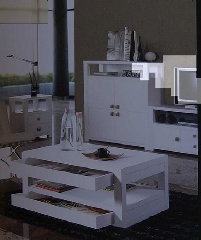}
	\includegraphics[width = 0.4\textwidth, height = 0.15\textheight]{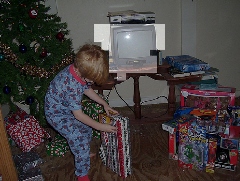}
	\caption{\textbf{TV Monitor} class localization examples from the PASCAL VOC dataset (test set).}
	\label{fig:tv}
\end{figure}

This section contains some additional figures to supplement the contents of the paper.  All details about the figures are included in their captions.
\begin{itemize}
  \item \textbf{Figure~\ref{fig:spamdetailed}} demonstrated the forward and backward pass through the SPAM pooling layer for an example input/gradient.
  \item \textbf{Figure~\ref{fig:bb}} is provided to highlight the differences in the backprojected areas between CAMs trained by global max pooling (GMP), global average pooling (GAP) and our spatial pyramid averaged max (SPAM) pooling methods.
  \item \textbf{Figure~\ref{fig:sovonexamples}} shows bird localization examples on a weakly labelled dataset of CCTV images from a nature reserve.
  \item \textbf{Figures~\ref{fig:air}--\ref{fig:tv}} show localization examples on images from the PASCAL VOC dataset (test set).
\end{itemize}

\clearpage

\clearpage

\clearpage

\end{document}